# "Hang in there": Lexical and visual analysis to identify posts warranting empathetic responses

**Mimansa Jaiswal[†], Sairam Tabibu[*], Erik Cambria[§]**

[†] Institute of Engineering and Technology, Indore, India

[*] Indian Institute of Technology, Varanasi, India

[§] Nanyang Technological University, Singapore

**Abstract**

Social media, in the past few years has risen as a platform where people express and share personal incidences about abuse, violence and mental health issues. There is a need to pinpoint such posts and learn the kind of response expected. For this purpose, we understand the sentiment that a personal story elicits on different posts present on different social media sites, on the topics of abuse or mental health. In this paper, we propose a method supported by hand-crafted features to judge if the discourse or statement requires an empathetic response. The model is trained upon posts from various web-pages and corresponding comments, on both the captions and the images. We were able to obtain 80% accuracy in tagging posts requiring empathetic responses.

## Introduction

Artificial companions, though a hype are still pretty much limited in their capabilities as support systems. They are unable to establish meaningful relationships with users, widely used just as personal assistants that executes tasks rather than to talk to or communicate with. It is widely believed that a key aspect of establishing such rapport relies on empathy which is often seen as the basis of social cooperation and pro-social behavior.

Empathy is often defined as the verbal or non-verbal gestures that evoke a sense of understanding of others' state of mind in a particular situation. Empathy encompasses several human interaction abilities, especially those that require the competence to reconstruct other person's words or actions and their perceived consequences.

Previous research has widely shown that agents without empathy are less preferred as compared to those who are empathetic, the latter being considered caring and likeable [1, 2]. Empathy involves perspective taking, developing sensitivity to the other's affective state and communication of a feeling of care [3]. As such, empathy is often related to helping behavior and friendship: people tend to feel more empathy for friends than for strangers, and vice-a-versa. With the penetration of voice assistants such as Siri, Google Voice and Cortana, and more people turning to in times of health crisis, it is alarming to see the apathy with which they respond. Research shows that how someone responds to you when you are feeling low and disclosing a private crisis, can affect how you act and feel.

General health disclosures can be divided primarily into four categories:

(a) **Mental Health:** Issues related to stress, depression, feeling low, restless

  "I am feeling low. I want to commit suicide"

(b) **Violence:** General acts of abusive behavior such as domestic violence, rape

  "Today, I was raped"

(c) **Needing support:** Posts about losing a family member, tragedies which are temporal and not clinical

  "I lost Pluto today. He was the sweetest dog I had ever known."

(d) **Physical Health:** Cases of physical discomfort such as sweating, pain, heart attack

  "Please help me. I think I am having a heart attack."

Such cases of disclosures have been mapped and tracked through psychology. While there are many applications that address the health point of view, mental health remains a distasteful subject, often hidden and sidelined as a case of continuous worry, or laziness or similar sense of disapproval towards the victim. We refer to disclosures in the category 1, 2 and 3 as "*empathy-seekers*" thereafter.

The main contributions of the paper are as follows: (1) We propose a novel way to approach binary classification of empathy seekers. (2) We propose a generalized list of features that could work on different categories of posts. (3) We develop a standard corpus for empathy seekers (using search queries such as *soul-stirring, depression* from web-pages explained in the later categories)

The rest of the paper is structured as follows: Section II lists related works in empathy and affect; Section III presents the dataset development technique; Section IV provides the proposed method; Section V elucidates the experiments and results and finally, Section VI concludes the paper and offers pointers for future work.

## Related Work

Prior research in psychology has examined the role of support from peers and society in combating mental health issues such as depression [26]. It is observed that nature of social support and individual's perception of support is often important and indispensable to achieve a timely recovery. Exploring research on social media, an increasing amount of work has shown that many people use the system to communicate around different health concerns [27].

Because many people go to social media to discuss and display events from their personal life, it is useful to examine how this kind of disclosure should be (a) perceived (b) marked (c) responded to. This information is especially helpful in case of mental health issues where the problem is socially stigmatic.

Because our work forms a part of psycholinguistics, it has been demonstrated that the use of linguistic patterns can reveal important social and psychological aspects of an individual [16]. Previous works have dealt with use similar psycholinguistic cues to measure a specific area of health issue such as depression [25], suicide [29] and bullying as well [28]. As far as we know, not any major success has been achieved in finding a model that fits across all categories, which is what usually social media platforms deal with. Previous works have studied discourse on reddit, which substantially marks the subreddits as decisive groups in disclosure, whereas tweets or facebook posts don't have such an ability.

In the area of empathetic response or social support, there are some primary categories of work being performed. Some of the studies base their dataset on professionally trained psychologist responses, and measure their success [30, 31]. Other works perform their evaluation on forums that do not involve content other than support messages, and hence classification of message type, in that case is a primary task [25], and we hope to incorporate this into our future tasks. We have therefore refrained from obtaining our dataset from specialized forums, to be able to weed out the apathetic responses as well. Many studies cater to empathetic/non-empathetic judgement with the aid of audio transcripts [31] of interviewing dataset which we do not consider as a modal in our study, for it is infrequently available in case of social-media disclosures.

## Dataset Development

The dataset was developed by sourcing images and captions with their respective response from various social media websites, namely tumblr, Facebook, Instagram and Buzzfeed. We skip twitter, which has been the natural choice for all social media analysis research, for its tendency to incline towards textual posts more than a combination of visual and textual, both. Also, on twitter, the distribution among retweets with response lies at a dismal 20%.

The main of dataset can be described as follows:

1) Develop a dataset which can be used to mark context that warrant an empathetic response.

2) Develop a dataset which can be used to identify empathetic and non-empathetic responses.

We utilize the Facebook API for storing posts from "Humans of New York" social page. This page posts stories from various topics such as war, refugee procedure, Syria bombing, wherein people disclose a part of their personal story. The story is usually accompanied by a picture which usually relates to the caption. The same API was then used to retrieve the top 3 corresponding empathetic comments from the post. Some stories on the page followed a distributed format where the story was spread out across various posts. For our purpose, we have merged those stories later, while keeping the original spread intact, to be used later.

The second set of data is collected from image sharing websites such as Instagram and tumblr. The usual format of a post on such websites is a caption, usually followed by one or several hashtags and then the responses. Because of the unavailability of a stable API, we hand-curate this dataset.

Lastly, we scrape images from social websites such as Buzzfeed and other listicles, which, though usually stated as click-bait, often echo the sentiments of the common public. Because the listicles are usually a group of images, with common comments, we presently copy these comments, corresponding to all these images for response testing.

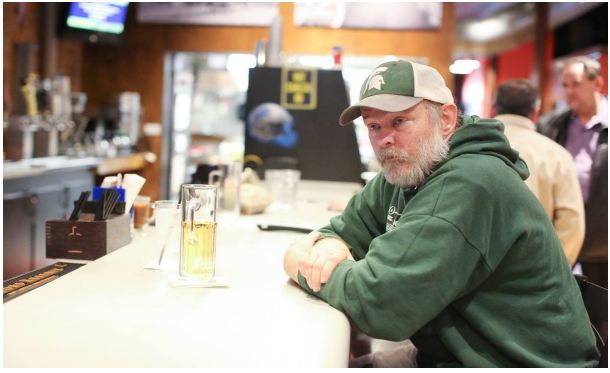

"If I think back, I get depressed. If I think ahead, I get afraid."

Response 1:
I feel you. Especially the thinking ahead... Anxiety and depression are no fun and I hope you find a way to overcome them and be happy!

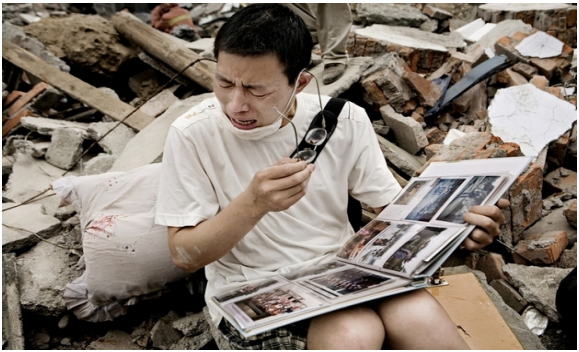

A survivor finds a photo album and cherished memories in the remains of his home after an earthquake in Sichuan, China.

Response 1: Such a beautiful thing..
Gives me a little more faith in the world.

*Figure 1 Sample images from Humans of New York and Buzzfeed that form a part of the database*

The comments, because they were hand-scraped, were also marked for gender, time duration from the post and the likes they received from the social media community. Our final database comprised of 1000 context-response pairs of positive examples that we use for this study. Because, privacy is a huge issue in social-media related research, the dataset is anonymized and stripped off of data that could be used for reverse-identification.

We add negative examples to our dataset by sourcing images related to happy events such as festivals, and by using search queries such as food, education and technology. We thoroughly check these added negative examples to verify our assumption that they do not seek an empathetic response. To add non-empathetic responses, we ask ten people to reply to the post, as if they were trying to belittle the author or to hold them as a suspect. We also study comments from various sources, especially for mental health issues that were mentioned as demeaning and vile and added them to corresponding posts. For the purpose of keeping our dataset unbiased, we thereby use inter-annotator agreement between four annotators to decide, if the post was marked correctly, i.e. ES/NES (Empathy-seeker or Non-Empathy seeker) and ER/NER (Empathetic response or Non-empathetic response). We also try to weed out malicious and spam content from our dataset, by removal of common phrases and rebuttals that often occur on social media posts, especially on sites such as Buzzfeed.

The final dataset has the following distribution of context-response pairs: 330 of mental health issues, 283 of violence related issues and the remaining belonged to those requiring support.

## Proposed Method

We model the task of empathy-seeker detection as a supervised classification problem in which each post is either classified as empathy-seeking or non-empathy seeking. We use six sets of lexical features and three sets of visual features to build our model. In the following subsections, we detail the features used and the classifiers that have been tried and compared.

### Verbal Features

The verbal/textual features are used for two purposes: to classify the post as emotion seeker and to judge whether a response is empathetic or not.

### Baseline Features

n-grams have been known as the best task-independent features for textual classification [4]. Therefore, we choose n-grams as the baseline feature. We retrieved word n-grams, usually called as bag of words as bi-grams and tri-grams and skip-grams (bi-grams) which after tf-idf transformation form our corpus. We filtered all the n-grams whose frequency was less than five, in order to use only those n-grams that were essential to our model. This set of feature would henceforth be called as baseline. These n-gram features are also used to identify temporal features such as ***today, weeks** etc* which are then used to identify the posts falling into the temporal issue category from the three categories we mentioned above. These temporal features have been known to be good linguistic attribute in identifying self-disclosure posts [25].

### Lexical Features

To model sentiment, we used emotional information from SenticNet [5], a concept-level knowledge base for sentiment analysis that provides both semantic and affec-tive information associated to words and multiword expressions by means of commonsense computing [6, 7] and sentic computing [8]. SenticNet has been shown to model emotions such as satire [10], deception [11] and mood [12] appropriately in previous research tasks, and hence we believe that it would present an appropriate representation.

**Sentiment Amplification**

As a general trend, it can be observed that almost all empathetic responses on social media make use of smileys, or specific punctuations. The use of quotation (" ") [9] has been mentioned as an indication of inverse sentiment. Sentiment amplifiers are elements that draw attention to the sentiment conveyed in a statement by either intensifying the sentiment value or by negating it. They have been used successfully to model satirical texts [10], which is another form of emotion expression. Phrases such as "Oh Please!" and acronyms, were added to the feature list. The presence or absence of an amplifier is used to form the feature vector.

**Speech act Features**

A speech act has a performative function in the context of language and communication, i.e. it performs the function of apology, appreciation, gratitude etc. [13]. In our study, we use 7 kind of speech act features, as stated: apology, appreciation, response acknowledgment, opinioned response, non-opinioned response, gratitude, other.

We build a speech oriented classifier from SPAAC [14] using the above-mentioned features as input. The classifier built had an accuracy of 72%, which we then use to find the speech act distribution over our corpus. Because the feature takes only one sentence at a time, to map a whole context or a response, we normalize the probability of occurrence of each speech act over the number of statements present. This is especially useful for posts that have long captions or those responses that are of more than a few lines.

**Literary Device features**

Because many of the expressions are metaphorical/indirect in nature or convey a sense of urgency or a feeling of helplessness, we use the following literary features as well.

**Hyperbole:** Hyperbole is referred to as statements that tend to exaggerate the actual sentiment. This is usually mapped by occurrence of multiple positive or negative words consecutively [15].

**Imagery:** These are the words that create a visual understanding in mind of the reader. For example, "He took me to a close dark cabin", would be an example of imagery.

**Psycholinguistic Features**

To extract psycholinguistic features, we utilize the Linguistic Word Count (LIWC) [16], which is a knowledge based system that has been developed upon in the past decade. The utility of such features has been studied in various areas such as personality, age, deception, health. The types of LIWC features we use are:

- General: word count, average words, word length
- Psycholinguistic: affect, cognition
- Personal concerns: work, achievement, home

**Visual Features**

Most of the websites allow addition of images along with posts. These images are often personal images, though they could also be images with text or commonly used stock images. We use visual features to model personal images and know which of these warrant an empathetic response.

**Facial presence**

The first feature we use if based upon whether there is a presence of a face in the image or not. The model is run on both, the data from pages other than Humans of New York, and that combined. This distinction is maintained because the images on the page are professionally captured and don't reflect the other variables into account (such as filters, angles etc.) The feature vector models the presence of image, and if present, how many of them were there. We believe that self-focus extends to photographs too [17], while measuring isolation. We use an elementary face detection script based on an open source demonstration.

**Gaze and facial sentiment**

The second set of feature took into account the gaze, if face was present, whether the participant was directly looking into the camera or away, and the facial angle from the vertical line. We also use OpenFace to measure Facial Action units and classify the sentiment projected by the face in the image, or average of the sentiments projected by the faces in the image. The sentiment probability was measured across 8 categories namely, anger, contempt, disgust, fear, happiness, neutral, sadness, surprise. These three criteria were used to correlate introvert nature, social anxiety and isolation and hence form a part of our feature vector.

**Hue and color**

We take into account the image properties namely Hue, Saturation and Value. These three-color properties are commonly used in image analysis. Hue describes the coloring on light spectrum ranging from red to blue, where low value lies on left and the high values lie on the right. Saturation refers to the radiance of the image, whereas value refers to image brightness, where lower the score, darker the image.

It has been observed that the happy individuals prefer vivid colors, while those feeling low or in need of support prefer darker colors [18, 19]. We calculate pixel level averages to obtain HSV for our feature set, which have been previously noted as satisfactory markers for mental health issues [20].

## Experiments

We used three different classification methods to test the accuracy of our features, namely Logistic Regression (LR), Random forest (RF) and an ensemble of both of them. We perform an ensemble of LR and RF based on majority voting scheme. We use these two classifiers because they have the minimum relation amongst them i.e. one models linear features while the other models non-linear ones.

We model ensemble classifier as follows:

*Ensemble_Classifier = w1\*LR + w2\*RF*

where *w1+w2=1.0* and *w1, w2* belong to {0.1, 0.9}.

We iterated over all possible combinations of w1 and w2 for the minimum cross entropy and settled upon 0.7 and 0.3 respectively. Individually, logistic regression produced the best result with an accuracy of over 76% with 99% confidence, while random classifiers averaged over 70% overall, probably due to overfitting on images. But a simple ensemble shoots the prediction accuracy of our model significantly, raising it up to 80.2% for overall classification and also 84% in some cases.

Table 1 represents the f-scores of our model using different classification techniques with different feature sets on partitioned datasets.

| Features | LR | RF | LR+RF |
|---|---|---|---|
| **Empathy Seekers Classification** | | | |
| **Verbal** | | | |
| BF | 65% | 60% | 70.2% |
| BF+LF+SA | 66.23% | 62.11% | 73.03% |
| BF+LF+SA +LD | 65.34% | 63.72% | 73.59% |
| BF+LF+SA+SF+LD+PF [a] | 69.87% | 65.93% | 76.24% |
| **Visual** | | | |
| FP | 58% | 50.1% | 70% |
| FP+ HSV | 64% | 61.2% | 73% |
| FP+GFS | 66% | 61% | 73.2% |
| FP+GFS+HSV [b] | 68% | 63.2% | 74.33% |
| **Verbal + Visual ([a]+[b])** | | | |
| Mental Health (MH) | 73.3% | 69.40% | 80% |
| Temporal Support (TS) | 76.77% | 69.78% | 84% |
| Violence and Abuse (VA) | 70.23% | 65.18% | 76.6% |
| MH + TS + VA | 73.43% | 68.12% | 80.2% |
| | | | |
| **Empathetic Response Classification** (Only verbal features) | | | |
| BF | 66.13% | 63.2% | 73% |
| BF+LF+SA | 68.2% | 64.55% | 73.33% |
| BF+LF+SA +LD | 69.87% | 66.71% | 75.6% |
| BF+LF+SA+SF+LD+PF [a] | 72.16% | 69.33% | 78.9% |

## Conclusion

We have proposed a method for identifying posts that require empathetic response. We have also tried to classify responses as empathetic or non-empathetic using a suite of classifiers. We believe this is one of the preliminary works that attempts to incorporate both text and images, without the aid of speech, which has been known as the primary method for support detection in various affect based conversational models.

Our model performs significantly well on classifying empathetic and non-empathetic responses, the f-scores averages to 79%, which, though cannot be compared to a benchmark due to lack of work in this area in the same context, beats the score of empathetic classification in call-center context of 70% [24]. We believe this could be an important aspect in marking spam or hurtful responses or those that violate *Be Nice, Be Respectful* policy in social media forums. Identifying types of social support in reddit commentary involved characterizing the content of comments into thematic clusters which we aim to explore with LDA in our future work.

We observe that ensemble classifiers perform the best and our use of gaze and HSV values in images combine with verbal features to give a superior performance. We believe that this performance would be enhanced if we took into account the photos that do not contain faces, but rather text or are stock images.

In future, we aim to deploy neural network for learning features other than our hand-crafted ones for they have been found to reduce the size of feature vector immensely [23, 24]. Also, the descriptors we derived in this paper could be used to develop user-adaptive response systems, especially in conversational agents, by recognizing if a context warrants support and replying accordingly. We aim to experiment with Dual Encoded LSTM for dialogue generation for the above-mentioned purpose.

## References

1. S. Brave, C. Nass, and K. Hutchinson, "Computers that care: investigating the effects of orientation of emotion exhibited by an embodied computer agent", Int. J. Human Computer. Stud., 62(2):161–178, 2005.

2. J. Paiva, D. Dias, R. Sobral, P. Aylett et al., "Caring for agents and agents that care: Building empathic relations with synthetic agents", In AAMAS '04: Proceedings of the Third International Joint Conference on Autonomous

Agents and Multiagent Systems, pages 194–201, Washington, DC, USA, IEEE Computer Society, 2004.

3. P. Goldstein and G. Y. Michaels. Empathy: development, training, and consequences. Arnold P. Goldstein, Gerald Y. Michaels. New American Library, 1985.

4. J. Furnkranz, "A study using n-gram features for text categorization," Austrian Research Institute for Artifical Intelligence, vol. 3, no. 1998, pp. 1–10, 1998.

5. E. Cambria, S. Poria, R. Bajpai, and B. Schuller, "SenticNet 4: A semantic resource for sentiment analysis based on conceptual primitives", in: proceedings of COLING, Osaka, 2016.

6. E. Cambria, A. Hussain, C. Havasi, and C. Eckl, "Common sense computing: From the society of mind to digital intuition and beyond," in Biometric ID Management and Multimodal Communication (J. Fierrez, J. Ortega, A. Esposito, A. Drygajlo, and M. Faundez-Zanuy, eds.), vol. 5707 of Lecture Notes in Computer Science, pp. 252–259, Berlin Heidelberg: Springer, 2009.

7. E. Cambria, T. Mazzocco, A. Hussain, and C. Eckl, "Sentic medoids: Organizing affective common sense knowledge in a multi-dimensional vector space. In: Advances in Neural Networks", (D. Liu, H. Zhang, M. Polycarpou, C. Alippi, and H. He, eds.), vol. 6677 of Lecture Notes in Computer Science, (Berlin), pp. 601–610, Springer-Verlag, 2011.

8. E. Cambria, A. Hussain, C. Havasi, and C. Eckl, "Sentic computing: Exploitation of common sense for the development of emotion-sensitive systems," in Development of Multimodal Interfaces: Active Listening and Synchrony (A. Esposito, N. Campbell, C. Vogel, A. Hussain, and A. Nijholt, eds.), Lecture Notes in Computer Science, pp. 148–156, Berlin: Springer, 2010.

9. K. Buschmeier, P. Cimiano, and R. Klinger, "An impact analysis of features in a classification approach to irony detection in product reviews," in Proceedings of the 5th Workshop on Computational Approaches to Subjectivity, Sentiment and Social Media Analysis, (Baltimore, Maryland), pp. 42–49, Association for Computational Linguistics, June 2014.

10. A.N. Reganti, T. Maheshwari, U. Kumar, A. Das and R. Bajpai, "Modelling Satire in English Text for Automatic Detection", in Proceedings of Workshop on SENTIRE, (Barcelona), ICDM, December 2016.

11. M. Jaiswal, S. Tabibu, R. Bajpai, "The truth and nothing but the truth: Multimodal analysis for deception detection", in Proceedings of Workshop on SENTIRE, (Barcelona), ICDM, December 2016.

12. F. Alam, F. Celli, Evgeny A. Stepanov et al., "The Social Mood of News: Self-reported Annotations to Design Automatic Mood Detection Systems", in proceedings of COLING, Osaka (2016)

13. R. E. Sanders, "Dan sperber and deirdre wilson, relevance: Communication and cognition, oxford: Basil blackwell, 1986. pp. 265.," Language in Society, vol. 17, no. 04, pp. 604–609, 1988.

14. G. Leech and M. Weisser, "Generic speech act annotation for taskoriented dialogues," in proceedings of the 2003 Corpus Linguistics Conference, pp. 441Y446. Centre for Computer Corpus Research on Language Technical Papers, Lancaster University, 2003.

15. R. Gibbs and H. Colston, "Irony in Language and Thought: A Cognitive Science Reader", Lawrence Erlbaum Associates, 2007.

16. J.W. Pennebaker, M.E. Francis, and R.J. Booth, "Linguistic inquiry and word count: Liwc 2001", Mahway: Lawrence Erlbaum Associates, 71, 2001.

17. S. Rude, EM Gortner, JW Pennebaker, "Language use of depressed and depression-vulnerable college students. Cogn Emotion", 18(8): 1121–1133, 2004.

18. C.J. Boyatzis, R. Varghese, "Children's emotional associations with colors". J Genet Psychol 155(1): 77-85.

19. HR Carruthers, J. Morris, N. Tarrier, PJ Whorwell, "The Manchester Color Wheel: development of a novel way of identifying color choice and its validation in healthy, anxious and depressed individuals", BMC Med Res Methodol 10: 12, 2010.

20. A.G. Reece, C. Danforth, "Instagram photos reveal predictive markers of depression", pre-print: arXiv:1608.03282v2.

21. A. Severyn and A. Moschitti, "Twitter sentiment analysis with deep convolutional neural networks," in Proceedings of the 38th International ACM SIGIR Conference on Research and Development in Information Retrieval, ACM, 2015.

22. S. Poria, E. Cambria, and A. Gelbukh, "Deep convolutional neural network textual features and multiple kernel learning for utterance-level multimodal sentiment analysis," in EMNLP, 2015.

23. S. Poria, E. Cambria, and A. Gelbukh, "Aspect extraction for opinion mining with a deep convolutional neural network," Knowledge-Based Systems, vol. 108, 2016.

24. F. Alam, M. Danieli and G. Riccardi., "Can We Detect Speakers' Empathy? A Real-Life Case Study", in proceedings of COLING, IEEE International Conference Cognitive InfoCommunications, 2016.

25. M.D. Choudhary, S. De, "Mental Health Discourse on reddit: Self-disclosure, Social Support, and Anonymity", in proceedings of the Eight International AAAI Conference on Weblogs and Social Media, 2014.

26. LK George, DG Blazer, DC Hughes et al., "Social support and the outcome of major depression." The British J. of Psychiatry,154(4), 478-485, 1989.

27. M. Paul and M. Dredze, "You are what you tweet: analyzing Twitter for public health", in proceedings ICWSM, 2011.

28. J. Xu, X. Zhu and A. Bellmore, "Fast Learning for Sentiment Analysis on Bullying", in Proceedings of the First International Workshop on Issues of Sentiment Discovery and Opinion Mining, 2012.

29. J. Pestian, P. Matykiewicz, M. Linn-Gust et al. "Sentiment analysis of suicide notes: A shared task", in proceedings of Biomedical informatics insights, 2012.

30. T. Althoff, K. Clark, J. Leskovec, "Natural Language Processing for Mental Health: Large Scale Discourse Analysis of Counseling Conversations", pre-print: arXiv: 1605.04462v1

31. J. Gibson, B. Xiao, Z. Imel et al., "A Deep Learning Approach to Modeling Empathy in Addiction Counseling", in proceedings of Interspeech, 2016.